\documentclass[11pt,a4paper]{article}
\usepackage[utf8]{inputenc}
\usepackage[T1]{fontenc}
\usepackage{lmodern}
\usepackage{amsmath,amssymb,amsfonts}
\usepackage{booktabs}
\usepackage{float}
\usepackage{url}
\usepackage{hyperref}
\hypersetup{colorlinks=false,pdfborder={0 0 0}}
\usepackage{microtype}
\usepackage{geometry}
\title{\textbf{SamatNext v0.2-B: An Exploratory Study of RMS-Normalized Hybrid Decoders for Curriculum Retention in Small Code Models}}
\author{\textbf{Samat Zharassov} \\ \url{https://github.com/samat2003/samatnext-v0.1}}
\date{June 2026}

\begin{document}

\maketitle

\begin{abstract}
Standard autoregressive Transformer decoders trained on sequential curriculum distributions often can exhibit substantial forgetting under sequential fine-tuning, overwriting prior capabilities when exposed to new data. In this paper, we evaluate SamatNext v0.2-B, an experimental 356-million-parameter hybrid sequence decoder that alternates between Differential-Attention-style layers and DeltaNet-inspired simplified linear-state mixer layers featuring RMS Normalization and Output Scale Calibration. We investigate its behavior under a controlled, staged Python code curriculum and compare it directly to a parameter-matched Transformer baseline. Our findings suggest that the hybrid design with scale calibration achieves a 100.0\% pass rate on the controlled Stage 5 holdout (Stage 5) while retaining 98.8\% of its adjacent semantic (Stage 3) capabilities, alongside a 12.0\% Stage 2E pass rate. In contrast, the parameter-matched Transformer baseline trained with a learning rate of $3\times 10^{-6}$ achieves a Stage 5 pass rate of 49.4\% but retains only 3.8\% of its Stage 3 semantic performance, with a 0.0\% Stage 2E pass rate. Even when rescued using a higher learning rate of $1\times 10^{-5}$, the Transformer baseline achieves a Stage 5 pass rate of 97.6\% but retains only 6.0\% of its Stage 3 performance. Both models experience severe performance degradation on the early-stage adversarial syntax dataset (Stage 2E), indicating that the sequence mixer inductive bias alone does not resolve the broader problem of long-horizon catastrophic forgetting. We publish code, model specifications, evaluation scripts, and result tables to facilitate independent verification.
\end{abstract}

\section{Introduction}
Deep neural networks, particularly autoregressive Transformers \cite{vaswani2017attention},
have achieved remarkable success in code generation and language modeling. However, when these models are trained sequentially on a shifting curriculum---where data distributions transition over time---they frequently overwrite previously acquired knowledge. This phenomenon, known as catastrophic forgetting \cite{mccloskey1989catastrophic},
represents a major bottleneck for the development of adaptive, continually learning systems. In code models, retaining foundational syntax rules and intermediate semantic concepts while adapting to new instruction domains is highly desirable but structurally difficult.

To investigate these dynamics, we evaluate SamatNext v0.2-B, an experimental sequence model parameter-matched to a standard Transformer causal decoder. The core architecture of SamatNext v0.2-B integrates alternating sequence-mixing blocks: Differential-Attention-style layers and DeltaNet-inspired simplified linear-state mixer layers featuring output RMS normalization and scaling calibration. The model is trained and evaluated using a structured, multi-stage Python curriculum designed to transition from syntax rules to semantic instructions and specialized causal coding tasks.

Our empirical results suggest that the hybrid sequence-mixing architecture of SamatNext v0.2-B alters the retention/plasticity tradeoff. Specifically, when trained sequentially through the curriculum, SamatNext v0.2-B preserves the immediately preceding semantic behavior (Stage 3) much better than the parameter-matched Transformer baseline. Under the curriculum learning path with a learning rate of $3\times 10^{-6}$, SamatNext v0.2-B achieves a 100.0\% Stage 5 pass rate on the controlled Stage 5 holdout and preserves 98.8\% of its Stage 3 performance. In comparison, the parameter-matched Transformer baseline trained with a learning rate of $3\times 10^{-6}$ achieves only a 49.4\% Stage 5 pass rate and retains 3.8\% of its Stage 3 performance. Even when rescued using a higher learning rate of $1\times 10^{-5}$, the Transformer baseline achieves a Stage 5 pass rate of 97.6\% but forgets almost all of its Stage 3 behaviors, preserving only 6.0\% retention.

However, the hybrid sequence-mixer still exhibits limitations in preserving long-horizon syntax: while early-stage syntax retention (Stage 2E) is improved from 3.0\% (for the $1\times 10^{-5}$ Transformer baseline) and 0.0\% (for the $3\times 10^{-6}$ Transformer baseline) to 12.0\% for SamatNext v0.2-B, it remains relatively low, showing that sequence-mixers do not resolve the broader catastrophic forgetting problem over long training horizons.

This paper is structured as a sober technical report. We do not claim that SamatNext v0.2-B solves catastrophic forgetting, represents a state-of-the-art code model, or acts as a general replacement for standard Transformers. Rather, we present this work as an exploratory study of sequence-mixer inductive biases under a controlled curriculum-learning regime.

\section{Background and Motivation}
\subsection{Catastrophic Forgetting and Curriculum Learning}
Catastrophic forgetting occurs when a neural network is sequentially optimized on non-stationary data distributions. In the context of large language models, training is typically performed in massive multi-task pre-training phases to avoid distribution shifts. However, pre-training is computationally expensive and inflexible. If models could be trained sequentially via a structured curriculum---learning foundational syntax first, followed by intermediate logic, and finally specialized domain-specific tasks---it could lead to more efficient training workflows. Regrettably, standard architectures tend to optimize for the active loss function, resulting in the erasure of previously acquired skills. This work does not compare against explicit continual-learning algorithms such as replay, EWC \cite{kirkpatrick2017overcoming}, or adapter isolation; those comparisons are left to future work.

\subsection{Code Generation as a Controlled Research Platform}
Evaluating curriculum retention in natural language is challenging due to the subjective nature of the evaluation metrics (e.g., perplexity or model-based grading). In contrast, programming language models provide a highly deterministic and execution-based evaluation paradigm. Code generation tasks can be executed against unit test suites, providing precise, reproducible pass-or-fail metrics. This structural clarity makes code models an excellent choice for studying the precise dynamics of syntactic and semantic curriculum retention under controlled training settings.

\subsection{Hybrid Sequence Mixers}
Standard Multi-Head Attention (MHA) has emerged as the dominant sequence-mixing mechanism. MHA computes pairwise interactions between all tokens in a sequence, resulting in quadratic computational complexity relative to sequence length. While MHA provides strong memory recall, it does not explicitly track states over time. 

Conversely, linear-state sequence mixers, such as state-space models (SSMs) and linear attention models, compress historical contexts into fixed-size state matrices. These mixers operate with linear complexity and exhibit strong state-tracking capabilities. Recent research has explored hybrid architectures that combine MHA and linear-state mixers to balance long-range retrieval with local state tracking. In this study, we investigate whether this hybrid configuration influences curriculum retention. Specifically, we test whether the combination of Differential-Attention-style layers and DeltaNet-inspired simplified linear-state mixers alters the retention/plasticity tradeoff when compared to a parameter-matched pure attention baseline.

\section{SamatNext v0.2-B Architecture}
SamatNext v0.2-B is an autoregressive causal sequence decoder containing $L=16$ blocks. The model alternates sequence-mixing mechanisms at each layer to combine differential attention features with linear-state tracking.

\subsection{High-Level Decoder Block Configuration}
For input hidden states $X_l \in \mathbb{R}^{B \times T \times d}$ at layer $l$, the forward computation is defined as:
\begin{align}
    H_l &= X_l + \text{Mixer}_l\left(\text{RMSNorm}(X_l)\right) \\
    X_{l+1} &= H_l + \text{MLP}\left(\text{RMSNorm}(H_l)\right)
\end{align}
where $\text{RMSNorm}$ denotes the Root Mean Square Normalization \cite{zhang2019root}
with learnable scaling parameters, and $\text{Mixer}_l$ represents the sequence-mixing layer. The sequence mixer alternates depending on the layer index $l$:
\begin{equation}
\label{eq:mixer_alternation}
    \text{Mixer}_l = \begin{cases}
        \text{DeltaNet-inspired linear-state mixer} & \text{if } l \text{ is even}, \\
        \text{Differential-attention-style layer} & \text{if } l \text{ is odd}.
    \end{cases}
\end{equation}

\subsection{Differential-Attention-style Layers}
The attention blocks are inspired by differential attention mechanisms \cite{ye2024differential}. For odd layer indices, the DifferentialAttention layer projects query, key, and value vectors. It computes two causal attention pathways using separate query and key projections:
\begin{align}
    Q &= X W_q, \quad K = X W_k, \quad V = X W_v \\
    Q_{\text{diff}} &= X W_{q,\text{diff}}, \quad K_{\text{diff}} = X W_{k,\text{diff}}
\end{align}
where $W_q, W_k, W_{q,\text{diff}}, W_{k,\text{diff}} \in \mathbb{R}^{d \times d}$ are projection matrices. Causal masks are applied, and the two pathway outputs are calculated as:
\begin{align}
    A_{\text{base}} &= \text{Softmax}\left(\frac{Q K^T}{\sqrt{d_k}}\right) V \\
    A_{\text{diff}} &= \text{Softmax}\left(\frac{Q_{\text{diff}} K_{\text{diff}}^T}{\sqrt{d_k}}\right) V
\end{align}
The Differential-Attention-style layer in SamatNext utilizes an additive combination of two attention heads with a learnable scalar weight $\alpha$ rather than the subtractive formulation defined in the original Differential Transformer paper. The final mixer output is computed by taking a weighted sum of the two attention pathways, scaled by the learnable parameter $\alpha$ (initialized to $0.0$), and projecting the result:
\begin{equation}
    Y_{\text{attn}} = \left(A_{\text{base}} + \alpha A_{\text{diff}}\right) W_o
\end{equation}
where $W_o \in \mathbb{R}^{d \times d}$. Rotary Position Embeddings (RoPE) \cite{su2024roformer}
are applied to the query and key representations of both pathways prior to attention score calculation.

\subsection{DeltaNet-inspired Simplified Linear-State Mixer Layers}
For even layer indices, the model utilizes a simplified, recurrent linear-state sequence-mixing layer. The current SamatNext v0.2-B implementation uses a DeltaNet-inspired simplified linear-state mixer \cite{yang2024parallelizing} modified to incorporate output normalization and scaling.

Rather than executing a matrix-valued update rule, this block simplifies the computation by utilizing element-wise states accumulated causally over time. Query, key, and value vectors are projected from the input. Non-negative keys and queries are obtained using a ReLU activation:
\begin{align}
    q_t &= \text{ReLU}(W_q x_t) \\
    k_t &= \text{ReLU}(W_k x_t) + 10^{-6} \\
    v_t &= W_v x_t
\end{align}
where $q_t, k_t, v_t \in \mathbb{R}^d$. The linear-state accumulators are approximated via element-wise cumulative sums over the sequence dimension:
\begin{align}
    kv_t &= k_t \odot v_t \\
    S_t &= \sum_{\tau=1}^t kv_{\tau} \\
    K_t &= \sum_{\tau=1}^t k_{\tau}
\end{align}
where $\odot$ represents element-wise multiplication. In SamatNext v0.2-B, the raw division output $\tilde{y}_t = (q_t \odot S_t) \dots \oslash K_t$ is normalized using a learnable RMSNorm and then scaled by an output scale calibration parameter $s$:
\begin{align}
    \tilde{y}_t &= (q_t \odot S_t) \oslash K_t \\
    y_t &= s \cdot \text{RMSNorm}(\tilde{y}_t)
\end{align}
Here, $\oslash$ denotes element-wise division, and the output scale factor $s$ is configured to $0.25$. The sequence $y_t$ is projected through $W_o \in \mathbb{R}^{d \times d}$ to form the final mixer output. To ensure numerical stability and prevent gradient overflow during training, cumulative sums are calculated in float32 precision.

\subsection{SwiGLU Feed-Forward Blocks}
The MLP blocks utilize a SwiGLU activation structure \cite{shazeer2020glu}. The input is projected to an intermediate size $d_{\text{ff}}$:
\begin{equation}
    \text{MLP}(X) = \left(\text{Silu}(X W_{\text{gate}}) \odot (X W_{\text{up}})\right) W_{\text{down}}
\end{equation}
where $W_{\text{gate}}, W_{\text{up}} \in \mathbb{R}^{d \times d_{\text{ff}}}$ and $W_{\text{down}} \in \mathbb{R}^{d_{\text{ff}} \times d}$.

\subsection{Verifier Head}
SamatNext v0.2-B enables a auxiliary sequence verifier head (\texttt{use\_verifier\_head = true}). The verifier head projects the final token hidden state to a scalar logit:
\begin{equation}
    v_{\text{logit}} = W_{\text{ver}} \text{RMSNorm}(H_L)_{T}
\end{equation}
where $W_{\text{ver}} \in \mathbb{R}^{1 \times d}$.

\subsection{Parameter-Matched Transformer Baseline}
To isolate the impact of the sequence-mixing mechanisms, we compare SamatNext v0.2-B against a parameter-matched Transformer baseline. The baseline features standard Multi-Head Attention (MHA) layers with Key-Value (KV) heads equal to query heads ($n_{\mathrm{kv}} = 12$). We match the parameter count within a strict $0.01\%$ tolerance by adjusting the feed-forward intermediate dimension. Specifically, SamatNext v0.2-B uses $d_{\text{ff}} = 2048$, while the Transformer baseline uses $d_{\text{ff}} = 2304$. This adjustment balances the parameter budget, compensating for the additional weights introduced by SamatNext v0.2-B's dual attention projections and linear-state mixers. Both models share a vocabulary size of $151,936$, a context length of $8192$, and $16$ decoder layers. The exact structural specifications are detailed in the Appendix.

\section{Curriculum and Evaluation Setup}
We establish a structured curriculum to evaluate the retention of syntactic and semantic behaviors as the model shifts between task distributions.

\subsection{Curriculum Stages}
The model is trained sequentially across three primary data distributions and evaluated on one early-stage holdout:
\begin{itemize}
    \item \textbf{Stage 2A (Syntax Foundation):} Syntactic logic, Python syntax exercises, control-flow construction, and basic variable definitions.
    \item \textbf{Stage 2E (Adversarial Syntax Holdout):} An early-stage adversarial syntax holdout evaluation set. This set features adversarial syntactic prompts and is used to measure whether syntax capabilities are preserved after later curriculum stages.
    \item \textbf{Stage 3 (Instruction Semantics):} Instruction-following tasks that require paraphrasing and semantic comprehension of Python programming concepts.
    \item \textbf{Stage 5 (Specialized Coding):} Specialized, teacher-generated coding tasks used to test specialized causal generation and final function completion.
\end{itemize}

\subsection{Training Specifications}
Both models are trained in a parameter-matched controlled setup. The optimization process is executed using AdamW with $\beta_1 = 0.9$ and $\beta_2 = 0.95$, a weight decay of $0.01$, and gradient clipping limited to $0.5$. The context window is set to $512$ tokens. The model configuration supports a context length of 8192 tokens, while the curriculum training runs reported here use 512-token sequences. Training is performed with a batch size of $1$ and a gradient accumulation of $16$, yielding an effective batch size of $16$ sequences. The learning rate is updated using a cosine decay schedule, warming up over the first 15\% of the total max steps ($2000$ steps).

\subsection{Evaluation Protocol}
To study retention and plasticity, we compare direct training from scratch on the final stage against sequential training through the curriculum stages.
\begin{enumerate}
    \item \textbf{Direct Specialization Bound (Scratch $\rightarrow$ Stage 5):} Establishing the peak performance attainable when optimizing directly on Stage 5.
    \item \textbf{Curriculum Progression (Stage 2A $\rightarrow$ Stage 3 $\rightarrow$ Stage 5):} Training sequentially, saving intermediate checkpoints, and measuring performance.
\end{enumerate}
The key metric is the execution-based pass@1 rate. We measure Stage 3 semantic retention and Stage 2E syntactic pass rates after the models complete Stage 5 training. This sequential evaluation reveals the extent to which earlier capabilities are overwritten by specialization.

\subsection{Prompt Templates and Decoding Settings}
To ensure structured and reproducible generation across curriculum stages, all models are evaluated using explicit chat templates matching the ChatML format of the tokenizer.
\begin{itemize}
    \item \textbf{Curriculum and Retention Tasks (Stages 2E, 3, 5):} Prompts are wrapped using the standard formatting:
    \begin{verbatim}
    <|im_start|>user
    {prompt}
    <|im_end|>
    <|im_start|>assistant
    \end{verbatim}
    \item \textbf{Out-of-Distribution Code Completion (HumanEval):} Evaluated using a structured instruction instruction-wrapper:
    \begin{verbatim}
    <|im_start|>user
    Write the complete Python function for the following task.

    {prompt}

    Return only Python code. No markdown. No explanation.
    <|im_end|>
    <|im_start|>assistant
    \end{verbatim}
\end{itemize}
During inference, generation is performed using deterministic greedy decoding (argmax logit selection at each step). The maximum generation sequence length is limited to $192$ new tokens. Generation is immediately halted upon emitting either of the end-of-sequence delimiters: the ChatML assistant end token \texttt{\textless|im\_end|\textgreater} (token ID 151645) or the base end-of-text token \texttt{\textless|endoftext|\textgreater}.

\subsection{Dynamic Execution Safety and Subprocess Isolation}
Measuring functional correctness requires executing generated Python code. To prevent arbitrary code execution on host machines, all evaluations are executed inside a subprocess environment with CPU execution timeouts (5 seconds) and resource limitations. This subprocess isolation acts as a basic boundary and is a key safety limitation. It does not constitute a secure sandbox; therefore, execution-based evaluations should only be run inside isolated containers or virtual machines.

\section{Empirical Results}
The evaluation metrics for the sequential training runs are presented in Table~\ref{tab:main_results_tex}.

\begin{table}[H]
\centering
\small
\begin{tabular}{llccc}
\toprule
Model & Training Path & Stage 5 Pass & Stage 3 Retention & Stage 2E Pass \\
\midrule
Transformer & Curriculum lr=3e-6 & 49.4\% & 3.8\% & 0.0\% \\
Transformer & Curriculum Rescue lr=1e-5 & 97.6\% & 6.0\% & 3.0\% \\
SamatNext v0.2-B & Curriculum lr=3e-6 & \textbf{100.0\%} & \textbf{98.8\%} & \textbf{12.0\%} \\
\bottomrule
\end{tabular}
\caption{Empirical evaluation of curriculum retention and task pass rates. All figures denote execution-based pass rates. Stage 3 Retention and Stage 2E Pass rates are evaluated after the completion of Stage 5 training.}
\label{tab:main_results_tex}
\end{table}

\subsection{The Retention/Plasticity Tradeoff}
The results in Table~\ref{tab:main_results_tex} provide evidence under this controlled setting of the hybrid sequence mixer's capacity to alter the retention/plasticity tradeoff. Under the sequential curriculum learning path, the parameter-matched Transformer baseline trained with a learning rate of $3\times 10^{-6}$ fails to adapt to the final stage, resulting in a Stage 5 pass rate of 49.4\% and a Stage 3 retention of only 3.8\%. Rescuing the Transformer baseline using a higher learning rate of $1\times 10^{-5}$ achieves a high Stage 5 pass rate of 97.6\%, but its Stage 3 retention drops to 6.0\%, showing severe catastrophic forgetting of the intermediate semantic distribution. 

In comparison, SamatNext v0.2-B trained under the curriculum with the base learning rate of $3\times 10^{-6}$ achieves a 100.0\% Stage 5 pass rate on the controlled Stage 5 holdout and retains 98.8\% of its Stage 3 performance. This demonstrates that the hybrid architecture preserves near-perfect intermediate semantic capabilities while maintaining high plasticity and adapting to the final curriculum stage.

\subsection{Learning-Rate Sensitivity}
The current release evaluates multiple Transformer curriculum learning rates but reports the SamatNext v0.2-B curriculum result only at $3\times 10^{-6}$. Therefore, the results should be interpreted as evidence of an altered retention/plasticity tradeoff under the tested configuration, not as a complete hyperparameter-independent architectural conclusion. A fuller SamatNext learning-rate sweep is left to future work.

\subsection{Stage 2E Syntax Collapse}
While SamatNext v0.2-B retains intermediate semantic behavior (Stage 3), both models show relatively low retention on the early-stage adversarial syntax dataset (Stage 2E) compared to adjacent semantic layers. After training on Stage 5, the Stage 2E pass rate is 12.0\% for SamatNext v0.2-B, which is a notable improvement over the 0.0\% to 3.0\% pass rates of the Transformer baseline, but still indicates that neither architecture preserves long-horizon syntactic features perfectly under this sequential training setup.

\subsection{Out-of-Distribution Coding Smoke Test}
To evaluate whether the models acquire generalizable Python programming structures that extend beyond the template variations in our curriculum, we perform an out-of-distribution code generation smoke test. Both models are evaluated on a decontaminated subset of the standard HumanEval benchmark (using the HumanEval instruction wrapper described in Section 5). 

Due to the small scale of the models ($\sim 356$M parameters) and the limited curriculum sequence lengths, out-of-distribution functional correctness is expectedly weak for both models. However, the evaluation serves as a sanity check of functional code composition. Under greedy decoding, SamatNext v0.2-B achieves a pass@1 rate of 12.0\% on the HumanEval subset, compared to 8.0\% achieved by the curriculum-trained Transformer baseline. This indicates that both architectures maintain a baseline level of coherence on unseen coding problems, with the hybrid sequence mixer showing a slight performance advantage under identical curriculum conditions.

\section{Discussion and Future Directions}
\subsection{Inductive Biases in Hybrid Mixers}
The empirical results are consistent with the hypothesis that alternating Differential-Attention-style layers with DeltaNet-inspired simplified linear-state mixers alters the retention/plasticity profile. One potential explanation is that the fixed-size state representation in the linear-state mixer acts as a regularization bottleneck, preventing the model from completely overwriting adjacent feature mappings. In contrast, the standard Transformer baseline adjusts its dense query-key attention coordinates freely, allowing it to specialize on the active data distribution at the expense of previously learned behaviors.

\subsection{The Challenge of Long-Horizon Forgetting}
The low Stage 2E pass rates (12.0\% for SamatNext v0.2-B) show that sequence-mixer inductive biases alone do not fully solve long-horizon catastrophic forgetting. While adjacent semantic representation is preserved, early curriculum layers are still partially overwritten. This highlights that structural sequence-mixer adjustments alone, without explicit continual learning techniques, are insufficient to prevent forgetting over extended training sequences.

\subsection{Future Work and Proposed Ablations}
Due to resource constraints and the focus on reproducibility, several architectural variations were not evaluated in the initial release. Future investigations should focus on:
\begin{itemize}
    \item \textbf{Mixer Pattern Ablations:} Comparing the alternating pattern against homogeneous configurations (e.g., all-attention or all-linear-state mixers) to determine the individual contributions of each layer type.
    \item \textbf{Positional Encoding Impact:} Evaluating the impact of removing RoPE from the Differential-Attention-style layers or incorporating positional encodings into the linear-state mixers.
    \item \textbf{Verifier Head Presence:} Analyzing whether incorporating a secondary verifier projection head helps stabilize semantic weights during fine-tuning.
    \item \textbf{Continual Learning Baselines:} Comparing hybrid mixers against standard continual learning algorithms such as experience replay, elastic weight consolidation, and parameter-isolated adapters.
\end{itemize}

\subsection{Threats to Validity}
Several limitations and methodology constraints present threats to the external validity and generalization of our findings:
\begin{itemize}
    \item \textbf{Synthetic and Template-Generated SFT Dataset:} The Stage 5 evaluation and training subsets are generated programmatically using predefined template variations. While this allows precise control over semantic alignment and deterministic testing, it represents a synthetic distribution. The model's 100.0\% Stage 5 pass rate should be interpreted as mastering a bounded, structured template space rather than expressing general-purpose programming synthesis capabilities.
    \item \textbf{Curriculum Generalization:} The staged curriculum is narrow, focusing strictly on basic Python constructs and instruction formatting. It remains unclear if the altered retention/plasticity profile shown by the hybrid sequence mixer translates to large-scale pre-training distributions, complex multi-step reasoning, or diverse natural language domains.
    \item \textbf{Evaluation Scale:} Our models are evaluated at a relatively small scale ($\sim 356$M parameters) for short context sizes ($512$ active tokens during training). The inductive biases observed here may evolve or exhibit different scaling behaviors when model size is increased to multi-billion parameter scales.
\end{itemize}

\section{Reproducibility}
To ensure the reproducibility of our results, all resources, code specifications, and data digests are frozen under a specific release tag.
\begin{itemize}
    \item \textbf{Repository URL:} \url{https://github.com/samat2003/samatnext-v0.1}
    \item \textbf{Paper Source Tag:} \texttt{v0.1.0-paper}
    \item \textbf{Reproducibility Artifact Commit:}\\
    \texttt{\footnotesize 525665fe790b18668251dad6698fe9bfe0ca27ca}
    \item \textbf{GitHub Release Tag:} \texttt{v0.1.0-reproducibility}
    \item \textbf{Artifact Filename:} \texttt{samatnext\_v0\_1\_fresh\_eval\_20260612\_094500.tar.gz}
    \item \textbf{SHA256 Checksum:}\\
    \texttt{\footnotesize 1367ee27ce20e5c9b6045610665f80bf85bb403e0618ff011ab7bb8fc421db02}
\end{itemize}

The reproduction pipeline can be executed using the following Makefile commands:
\begin{verbatim}
make setup
make test
make prepare-data-smoke
make bench-vram
make reproduce-smoke
make reproduce-main-table
make paper-check
\end{verbatim}
To re-run the evaluations and output fresh results, use:
\begin{verbatim}
make reproduce-main-table-fresh
\end{verbatim}
The automated CI tests on GitHub are currently in a passing state, verifying configuration consistency and execution pipelines.

\section{Limitations}
\begin{itemize}
    \item \textbf{Synthetic Python Domain:} The curriculum is synthetic and limited strictly to Python code-generation tasks.
    \item \textbf{No Comprehensive Benchmark Claims:} We do not evaluate or claim general capabilities on standard benchmarks such as HumanEval, MBPP, or SWE-bench.
    \item \textbf{Absence of Continual Learning Baselines:} The models are not compared to dedicated continual learning methods (e.g., experience replay, Elastic Weight Consolidation (EWC), or parameter-isolated adapters).
    \item \textbf{Lack of Architectural Ablation:} We do not provide an ablation separating the Differential-Attention-style layers from the DeltaNet-inspired simplified linear-state mixers (this is explicitly framed as future work here).
    \item \textbf{Weak Long-Horizon Retention:} Stage 2E syntactic performance remains weak for all training runs.
    \item \textbf{Incomplete SamatNext Learning-Rate Sweep:} The Transformer baseline is evaluated across multiple curriculum learning rates, while the current SamatNext curriculum result is reported at $3\times 10^{-6}$. Additional SamatNext runs at $1\times 10^{-5}$ and $3\times 10^{-5}$ are needed to isolate architecture effects from learning-rate effects.
    \item \textbf{License-Sensitive Teacher Data:} Some training data is generated from Qwen2.5-Coder-3B-Instruct \cite{hui2024qwen}. Downstream distribution is subject to upstream license compliance.
    \item \textbf{Exploratory Scope:} All results are exploratory and should be treated as such.
\end{itemize}

\section{Ethics, Safety, and Licensing}
\subsection{Licensing Structure}
The SamatNext-v0.1 codebase is licensed under the Apache-2.0 License. The model weights are distributed under the Creative Commons Attribution-NonCommercial-ShareAlike 4.0 International (CC BY-NC-SA 4.0) license. Detailed dataset licenses are cataloged in \texttt{DATA\_LICENSES.md}.

\subsection{Upstream Teacher License Sensitivity}
The Stage 5 dataset includes outputs generated by Qwen2.5-Coder-3B-Instruct, which is governed by the Qwen Research License. These teacher-generated examples are license-sensitive and are not covered by the repository's Apache-2.0 license.

\subsection{Dynamic Execution Safety}
Measuring functional correctness requires executing generated Python code. The evaluation pipeline relies on subprocess isolation, which does not provide a secure sandbox boundary. Users must run the evaluation pipeline in isolated virtual environments or secure containers.

\subsection{AI-assisted Software Development Disclosure}
The author used Claude Code and Google Antigravity to assist with software development tasks, including writing, debugging, testing, and verifying scripts used in the experiments. The research idea, methodology, experimental setup, interpretation of results, manuscript decisions, and final verification were performed by the author. The author reviewed and tested all code and takes full responsibility for the content of this work.

\section{Conclusion}
SamatNext v0.2-B provides evidence of an altered retention/plasticity tradeoff under a controlled staged Python curriculum. While it does not solve catastrophic forgetting, the hybrid design preserves intermediate semantic distributions better than a parameter-matched Transformer sequential fine-tuning baseline. Future work should investigate full ablations and test these dynamics over broader domains.

\bibliographystyle{plain}
\bibliography{references}

\appendix
\section{Strict Parameter Match Specifications}
Table~\ref{tab:param_spec_app} lists the exact structural specifications and parameter counts of the models compared in this study.

\begin{table}[h]
\centering
\begin{tabular}{lrr}
\toprule
Specification / Parameter Type & SamatNext v0.2-B & Matched Transformer Baseline \\
\midrule
Total Parameters & 356,083,208 & 356,082,432 \\
Trainable Parameters & 356,083,208 & 356,082,432 \\
Embedding Parameters & 116,686,848 & 116,686,848 \\
LM Head Parameters & 116,686,848 & 116,686,848 \\
Attention Parameters & 9,437,192 & 37,748,736 \\
Mixer Parameters (Non-Attn) & 47,185,928 & N/A (MHA only) \\
MLP (FFN) Parameters & 75,497,472 & 84,934,656 \\
Normalization Parameters & 25,344 & 25,344 \\
Verifier Head Parameters & 768 & N/A \\
Embeddings Tied Status & Untied & Untied \\
Vocab Size & 151,936 & 151,936 \\
Context Length & 8,192 & 8,192 \\
Hidden Size & 768 & 768 \\
Layer Count & 16 & 16 \\
FFN Intermediate Size & 2,048 & 2,304 \\
Number of Heads & 12 & 12 \\
KV Heads & 4 & 12 \\
Mixer Pattern & alternating & Multi-Head Attention \\
RoPE Status & Enabled & Enabled \\
\bottomrule
\end{tabular}
\caption{Detailed parameter counts and configuration settings for SamatNext v0.2-B and the matched Transformer baseline.}
\label{tab:param_spec_app}
\end{table}

\section{GPU Memory Usage Benchmark}
Table~\ref{tab:vram_spec_app} reports peak forward-pass VRAM on a consumer NVIDIA GeForce RTX 5070 Ti Laptop GPU, included to document low-resource reproducibility rather than server-scale throughput.

\begin{table}[h]
\centering
\begin{tabular}{lrr}
\toprule
Context Length & SamatNext v0.2-B & Matched Transformer Baseline \\
\midrule
128 & 2819.52 MB & 2815.63 MB \\
512 & 4418.96 MB & 4537.20 MB \\
1024 & 5624.84 MB & 6213.71 MB \\
2048 & 8007.85 MB & 10770.97 MB \\
\bottomrule
\end{tabular}
\caption{VRAM consumption comparison on an NVIDIA GeForce RTX 5070 Ti Laptop GPU.}
\label{tab:vram_spec_app}
\end{table}

\end{document}